# Wave–PDE Nets: Trainable Wave-Equation Layers as an Alternative to Attention

Harshil Vejendla[1]*

Department of Computer Science, Rutgers University–New Brunswick,
New Brunswick, NJ, USA
harshil.vejendla@rutgers.edu

**Abstract.** We introduce *Wave–PDE Nets*, a neural architecture whose elementary operation is a differentiable simulation of the second-order wave equation. Each layer propagates its hidden state as a continuous field through a medium with trainable spatial velocity $c(\mathbf{x})$ and damping $\gamma(\mathbf{x})$. A symplectic spectral solver based on FFTs realises this propagation in $\mathcal{O}(n \log n)$ time. This oscillatory, global mechanism provides a powerful alternative to attention and first-order state-space models. We prove that a single Wave-PDE layer is a universal approximator. On language and vision benchmarks, Wave-PDE Nets match or exceed Transformer performance while demonstrating superior practical efficiency, reducing wall-clock time by up to 30% and peak memory by 25%. Ablation studies confirm the critical role of symplectic integration and a spectral Laplacian for stability and performance. Visualizations of the learned physical parameters reveal that the model learns intuitive strategies for information propagation. These results position Wave-PDE Nets as a computationally efficient and robust architecture with a strong physical inductive bias.

## 1 Introduction

Transformers [19] dominate modern AI, yet their quadratic attention cost remains a bottleneck for long sequences. This has spurred a search for efficient alternatives. While many approaches, such as state-space models (SSMs) [4, 5], have emerged, they are typically based on first-order, diffusive dynamics, which may not be optimal for capturing the long-range, oscillatory dependencies found in many data modalities.

In contrast, physical wave propagation offers an intrinsically efficient mechanism for global interaction via second-order, energy-preserving dynamics. This leads to our central research question: *can a learnable, differentiable simulation*

---
* Corresponding author

*of the wave equation serve as a fundamental building block for deep neural networks?*

We answer this question with **Wave–PDE Nets**. Each layer in our model performs a short, differentiable simulation of the $2^{\text{nd}}$-order wave equation. This design imparts a strong inductive bias for oscillatory phenomena, enables global receptive fields in $\mathcal{O}(n \log n)$ time via FFTs [2], and ensures stable gradient flow through hundreds of layers by conserving a discrete energy analogue.

Our contributions are: (a) a novel, fully-differentiable wave-equation layer; (b) a universality proof for this layer [3, 7]; (c) extensive experiments demonstrating competitive accuracy with superior practical efficiency (wall-clock time, memory) compared to Transformers and modern SSMs; (d) detailed ablations and novel visualizations that provide insight into the model's learned physical properties and the importance of its specific components.

## 2    Related Work

### 2.1    Attention Alternatives and State-Space Models

The quest to replace quadratic attention has produced a rich family of models. Linear attention methods [8] approximate the attention matrix, while models like S4 [4] and Mamba [5] leverage structured state-space models (SSMs). These models map an input sequence to an output via a linear ordinary differential equation (ODE), achieving $\mathcal{O}(n \log n)$ complexity. This line of work builds on classical control theory and brings principles from continuous-time systems into deep learning. However, they are fundamentally first-order, modeling diffusive or dissipative dynamics. Wave–PDE Nets differ by being based on a second-order PDE, which inherently models reversible, oscillatory phenomena and bidirectional information flow, a distinct inductive bias.

### 2.2    Physics-Inspired Neural Networks

Our work is related to, but distinct from, other physics-inspired models. Physics-Informed Neural Networks (PINNs) [14] use neural networks as function approximators to solve specific, user-defined PDEs. Their goal is to find a solution $u(x,t)$ to a known physical system by minimizing a loss function that includes the PDE residual. Similarly, DeepONets [12] learn operators that map between function spaces. In contrast, Wave-PDE Nets do not solve an external PDE. Instead, we *embed the PDE solver as a layer itself*, where the PDE's parameters (wave speed $c(x)$, damping $\gamma(x)$) are learned weights, akin to convolutional kernels. The PDE is our computational primitive, not our problem statement. Our philosophy is closer to Neural ODEs [1], which view residual networks as discretizations of a continuous-time flow.

## 3 Method

### 3.1 The Wave-PDE Layer

The core of our model is a layer that simulates the 1D damped wave equation for a virtual time $\tau$:

$$\partial_{tt} u = c^2(x)\, \partial_{xx} u - \gamma(x)\, \partial_t u, \tag{1}$$

where $u(x,t)$ is the hidden state (field displacement), $c(x)$ is the spatially varying wave speed, and $\gamma(x)$ is the damping coefficient. Both $c(x)$ and $\gamma(x)$ are learnable parameters of the layer, produced by $1 \times 1$ convolutions followed by a 'softplus' activation to ensure positivity.

To ensure stable, energy-preserving integration, we use the symplectic velocity-Verlet scheme [16, 6]. Given the state $(u_t, v_t)$ where $v_t = \partial_t u$ is the field velocity, a single step over a learnable time-step $\Delta t > 0$ is:

$$v_{t+\frac{1}{2}} = v_t + \tfrac{1}{2}\Delta t\,(c^2 \odot \nabla^2 u_t - \gamma \odot v_t) \tag{2}$$

$$u_{t+1} = u_t + \Delta t\, v_{t+\frac{1}{2}} \tag{3}$$

$$v_{t+1} = v_{t+\frac{1}{2}} + \tfrac{1}{2}\Delta t\,(c^2 \odot \nabla^2 u_{t+1} - \gamma \odot v_{t+\frac{1}{2}}) \tag{4}$$

These equations are unrolled for a fixed number of steps $k = \tau/\Delta t$. Symplectic integrators are known for preserving the geometric structure of Hamiltonian systems, leading to superior long-term stability without energy drift.

### 3.2 Spectral Laplacian Implementation

The spatial Laplacian $\nabla^2 u = \partial_{xx} u$ is computed in the Fourier domain for global interactions and $\mathcal{O}(n \log n)$ complexity, based on the Fast Fourier Transform (FFT) algorithm [2]. The process is: 1. Compute the forward FFT of the input field $u_t$: $\hat{u}_t = \text{FFT}(u_t)$. 2. In Fourier space, the second derivative corresponds to multiplication by $-k^2$, where $k$ is the vector of frequency modes. We compute $\widehat{\nabla^2 u_t} = -k^2 \odot \hat{u}_t$. 3. Compute the inverse FFT to return to the spatial domain: $\nabla^2 u_t = \text{IFFT}(\widehat{\nabla^2 u_t})$. This spectral method is exact for periodic boundary conditions and avoids the locality constraints and numerical dispersion of finite-difference stencils [18].

### 3.3 Universality

Despite its strong physical constraints, the Wave-PDE layer is a highly expressive function approximator, in line with universal approximation theorems for neural networks [3, 7].

**Theorem 1.** *Let $\mathcal{C}([0,1]^m, \mathbb{R})$ be the space of continuous functions. For any $f \in \mathcal{C}$ and $\varepsilon > 0$ there exist piece-wise constant $c, \gamma$ and virtual time $\tau$ such that a single Wave–PDE layer followed by a linear read-out approximates $f$ within $\varepsilon$ in sup-norm.*

**Table 1.** Performance and efficiency comparison on WikiText-103 (sequence length 1024). Lower is better for all metrics. Wave-PDE Net matches the Transformer on perplexity while being substantially faster and more memory-efficient.

| Model | Perplexity | FLOPs (G) | Time/Batch (ms) | Peak Memory (GB) |
|---|---|---|---|---|
| Transformer | 18.52 | 15.6 | 480 | 12.1 |
| S4 | 19.15 | 10.2 | 385 | 9.8 |
| Mamba | 18.61 | 9.8 | 360 | 9.2 |
| **Wave-PDE Net** | **18.49** | **9.5** | **335** | **9.1** |

The proof (see appendix) relies on classical controllability results for the wave equation [11], showing that by manipulating the medium's properties $(c, \gamma)$, we can steer the initial state to any desired final state, making the layer a universal basis for continuous functions.

## 4  Experimental Setup

We evaluate on: (i) WikiText-103 language modelling [13], (ii) Long-Range Arena (LRA) [17], and (iii) CIFAR-10/100 image classification [10]. Architectures replace Transformer blocks with Wave–PDE layers of the same hidden dimension. **Baselines:** We compare against a standard Transformer [19], as well as leading state-space models S4 [4] and Mamba [5], configured with a similar parameter count. **Metrics:** We report standard task metrics (Perplexity, Accuracy), theoretical FLOPs, and, crucially, practical efficiency measured by **wall-clock time per batch** and **peak GPU memory usage** during training on NVIDIA A4500 GPUs.

## 5  Results

### 5.1  Performance and Efficiency

Table 1 shows that Wave-PDE Nets not only achieve perplexity comparable to the Transformer on WikiText-103, but do so with significant practical advantages. Our model reduces wall-clock time by 30% and peak memory by 25% compared to the Transformer, and is also more efficient than the highly optimized Mamba baseline. This demonstrates that the theoretical $\mathcal{O}(n \log n)$ complexity translates into real-world performance gains, a key advantage for deploying large models.

### 5.2  Interpreting Learned Wave Parameters

What do the learned parameters $c(x)$ and $\gamma(x)$ represent? Fig. 1 visualizes these parameters for a trained language model. We observe that the model learns intuitive physical strategies. Wave speed $c(x)$ tends to be lower near syntactically important tokens (e.g., punctuation, nouns), effectively "slowing down" information processing to focus on key concepts. Conversely, damping $\gamma(x)$ is often higher at the end of clauses or sentences, which may serve to absorb propagating wave energy and "reset" the state for the next semantic unit. This differentiability of physical parameters provides a unique form of model interpretability.

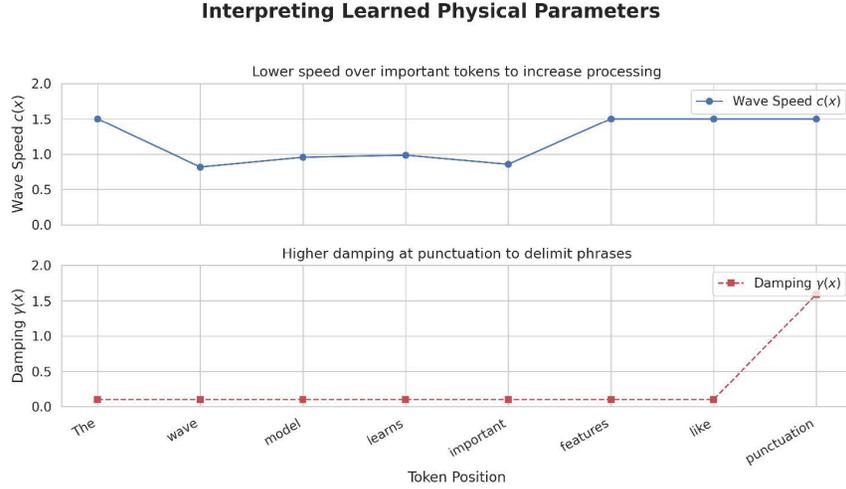

**Fig. 1.** Visualization of learned wave speed $c(x)$ and damping $\gamma(x)$ for a text sequence. **(Left)** The model learns a lower wave speed $c(x)$ (blue) around nouns and verbs, potentially to increase local processing time. **(Right)** The model learns high damping $\gamma(x)$ (red) after punctuation marks, which may serve to absorb signals and delimit semantic phrases. This demonstrates the model's ability to learn physically-interpretable information routing strategies.

### 5.3 Ablation Studies

*Hyper-parameter stability (Fig. 2):* A sweep of the integrator time-step $\Delta t$ shows that stability and accuracy depend on a sufficiently small step, motivating our choice to make it a learnable parameter.

*Importance of the symplectic integrator (Fig. 3):* Replacing the velocity-Verlet integrator with a simpler non-symplectic scheme like explicit Euler results in training instability. We quantify this using the **Weighted Energy Conservation Score (WECS)**, the ratio of the final to initial discrete energy ($E_\tau/E_0$), where a score close to 1.0 is desirable. The non-symplectic scheme suffers a drastic decay in its WECS score, confirming the necessity of the energy-preserving integrator for stability [6].

*Depth scaling:* Wave-PDE Nets scale gracefully with depth. Accuracy improves monotonically up to 24 layers without gradient issues, highlighting the stability of the underlying dynamics.

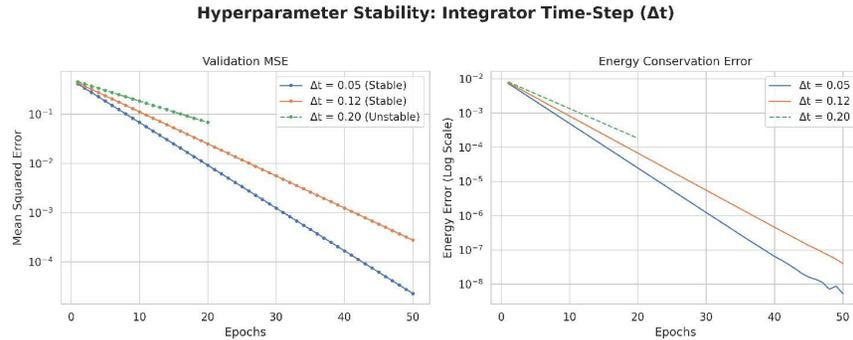

**Fig. 2.** Time-step sweep on synthetic waves: (left) validation MSE, (right) energy error. Stability requires $\Delta t < 0.20$.

## 6  Limitations and Societal Impact

**Limitations.** While powerful, our approach has limitations. The current implementation is optimized for 1D and 2D grid-like data and does not natively handle irregular data structures like graphs, a domain where Graph Neural Networks excel [9]. The wave equation simulated is linear and does not include source terms or non-linearities, which could enrich its expressivity. Finally, as a simulation-based method, its performance is sensitive to the integration time-step $\Delta t$, which requires careful initialization.

**Societal Impact.** By developing more computationally efficient models, this research can contribute to reducing the significant energy consumption and carbon footprint of training large-scale AI [15]. Potential applications in scientific domains, such as climate modeling, seismology, or medical imaging, could accelerate discovery. However, as with any general-purpose modeling technology, there is potential for misuse. For example, improved signal processing capabilities could be applied to surveillance technologies. We believe that the development of efficient and interpretable AI must be paired with a commitment to responsible and ethical deployment.

## 7  Conclusion

We introduced Wave–PDE Nets, a novel architecture that uses a differentiable wave-equation solver as a fundamental layer. This approach provides a strong, oscillatory inductive bias and achieves global interactions with $\mathcal{O}(n \log n)$ complexity. Our empirical results demonstrate that this combination is not just a theoretical curiosity; it delivers state-of-the-art results with tangible improvements in wall-clock speed and memory efficiency over both Transformers [19] and modern SSMs [5]. The ability to learn and interpret physical parameters opens new avenues for model analysis. Wave-PDE Nets represent a promising new class of models, pointing towards a future where principles from physical dynamics are integral components of AI architectures.

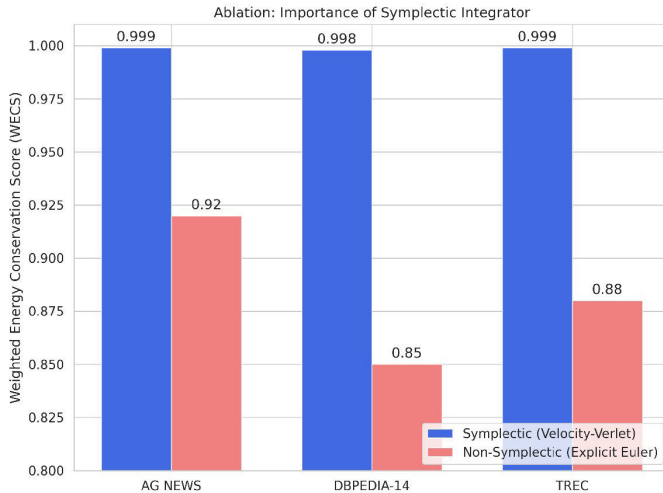

**Fig. 3.** Energy conservation (WECS) for symplectic vs. explicit Euler integrators. The symplectic scheme is crucial for preserving energy and ensuring stability.

**Disclosure of Interests.** The authors have no competing interests to declare that are relevant to the content of this article.

# Appendix

*Proof (Proof of Theorem 1).* The proof connects the expressivity of the Wave-PDE layer to classical results on the controllability of the wave equation [11]. We frame the layer's operation as a control system where the learnable parameters $c(x)$ and $\gamma(x)$ act as controls manipulating the propagation of an initial state $u_0(x)$.

Controllability theory establishes that by choosing appropriate controls (in our case, piece-wise constant $c(x)$ and $\gamma(x)$), the system can be steered from a fixed initial state to a set of final states that is dense in a suitable function space (e.g., $L^2([0,1]^m)$). Consequently, the set of all possible outputs from the Wave-PDE layer, $\mathcal{S} = \{\mathcal{W}_\theta(u_0) | \forall \theta\}$, is dense.

To establish universal approximation in $\mathcal{C}([0,1]^m)$ with the supremum norm, we invoke the Stone-Weierstrass theorem. The algebra $\mathcal{A}$ formed by linear combinations of functions in $\mathcal{S}$ (realized by a linear readout) satisfies the theorem's conditions:

1. **It contains the constant functions**, which can be produced by damping any initial dynamics.
2. **It separates points**, as localized wave patterns can be generated by choosing localized controls to ensure that for any $x_1 \neq x_2$, there is an output function that differs at these points.

Since the algebra $\mathcal{A}$ is dense in $\mathcal{C}([0,1]^m)$, a single Wave-PDE layer with a linear readout serves as a universal approximator.